\documentclass[10pt, a4paper]{article}
\newcommand{\method}{\textsc{RECAP}\xspace}

\usepackage{booktabs}
\usepackage{amsmath}
\usepackage{amssymb}
\usepackage{graphicx}
\usepackage{listings}
\usepackage{multirow}
\usepackage{subcaption}
\usepackage{xspace}
\usepackage{tcolorbox}
\tcbuselibrary{breakable,skins}
\usepackage{float}
\usepackage{tikz}
\usetikzlibrary{positioning,calc,arrows.meta,fit,backgrounds}
\usepackage{fontawesome5}
\usepackage[table]{xcolor}
\usepackage{pgfplots}
\pgfplotsset{compat=1.17}
\usepackage{enumitem}
\usepackage{microtype}
\usepackage{url}
\usepackage{array}
\usepackage{tabularx}

\usepackage{lrec2026} 
\title{\method: Transparent Inference-Time Emotion Alignment for Medical Dialogue Systems}

\name{Adarsh Srinivasan$^{\ast}$, Jacob Dineen$^{\ast}$, Muhammad Umar Afzal$^{\dagger}$, \\ {\bf \large Muhammad Uzair Sarfraz$^{\dagger}$, Irbaz B. Riaz$^{\dagger}$, Ben Zhou$^{\ast}$}}

\address{$^{\ast}$Arizona State University, Tempe, Arizona, USA \\
         \{asrini81, jdineen, xzhou202\}@asu.edu \\
         $^{\dagger}$Mayo Clinic, Tempe, Arizona, USA \\
         \{Afzal.MuhammadUmar, Sarfraz.MuhammadUzair, Riaz.Irbaz\}@mayo.edu \\}

\abstract{
Large language models in healthcare often produce emotionally flat or opaque responses, failing to provide the transparent reasoning required for clinical trust. We present \method{} (Reflect--Extract--Calibrate--Align--Produce), an inference-time framework grounded in cognitive appraisal theory that decomposes patient input into auditable, appraisal-theoretic stages without retraining. Across multiple benchmarks and models from 8B to 120B parameters, \method{} improves alignment with human judgments, with gains inversely proportional to model scale. Intermediate outputs further reveal that models systematically underweight relational factors such as social support. In blinded evaluations, oncology fellows rated \method{} responses significantly higher than baselines with 76--88\% win rates, demonstrating that principled prompting can enhance medical AI's emotional intelligence while maintaining the transparency required for clinical deployment.
\\ }

\begin{document}

\maketitleabstract

\section{Introduction}

Healthcare communication is fundamentally human-centered. When a patient receives difficult news or expresses anxiety, clinicians draw on training to recognize emotional cues, assess the patient's psychological state, and calibrate their response accordingly. This emotional attunement is central to building trust, supporting patient agency, and improving outcomes \citep{Chen2025PatientEmpathyCancerChatbots}. As conversational AI systems enter clinical settings, a fundamental question arises about how these systems process emotional content and whether we can make this reasoning transparent.

Current AI-assisted medical dialogue systems, while sophisticated in knowledge retrieval, leave emotional reasoning as a black box. A patient learning of a cancer diagnosis faces not only informational needs but profound emotional upheaval, including fear, uncertainty about treatment, and concerns about family. Two patients with identical diagnoses may require vastly different support depending on their sense of control, coping resources, and life circumstances \citep{Ortony1988OCC,Lazarus1991EmotionAdaptation}. Yet we cannot determine whether contemporary AI systems recognize these distinctions or simply pattern-match to empathetic-sounding language. In preliminary evaluations (Table~\ref{tab:emotional-intelligence-combined}), even state-of-the-art models consistently under-attend to affective dimensions that clinicians consider essential.

This opacity matters for both deployment and understanding. Prior work documents cases where AI responses felt ``cold'' or ``robotic'' despite being medically accurate \citep{Sorin2024JMIR,Armbruster2024}. Clinicians express reluctance to deploy systems whose emotional reasoning remains opaque, because they need to understand why an AI produces a particular response \citep{Chen2025PatientEmpathyCancerChatbots}. This need for interpretability distinguishes healthcare from domains where explainability is desirable but not essential. Beyond deployment concerns, this opacity limits our scientific understanding of how LLMs represent and reason about human emotion.

\begin{figure*}[t]
  \centering
  \includegraphics[width=0.88\textwidth,trim=6pt 6pt 6pt 6pt,clip]{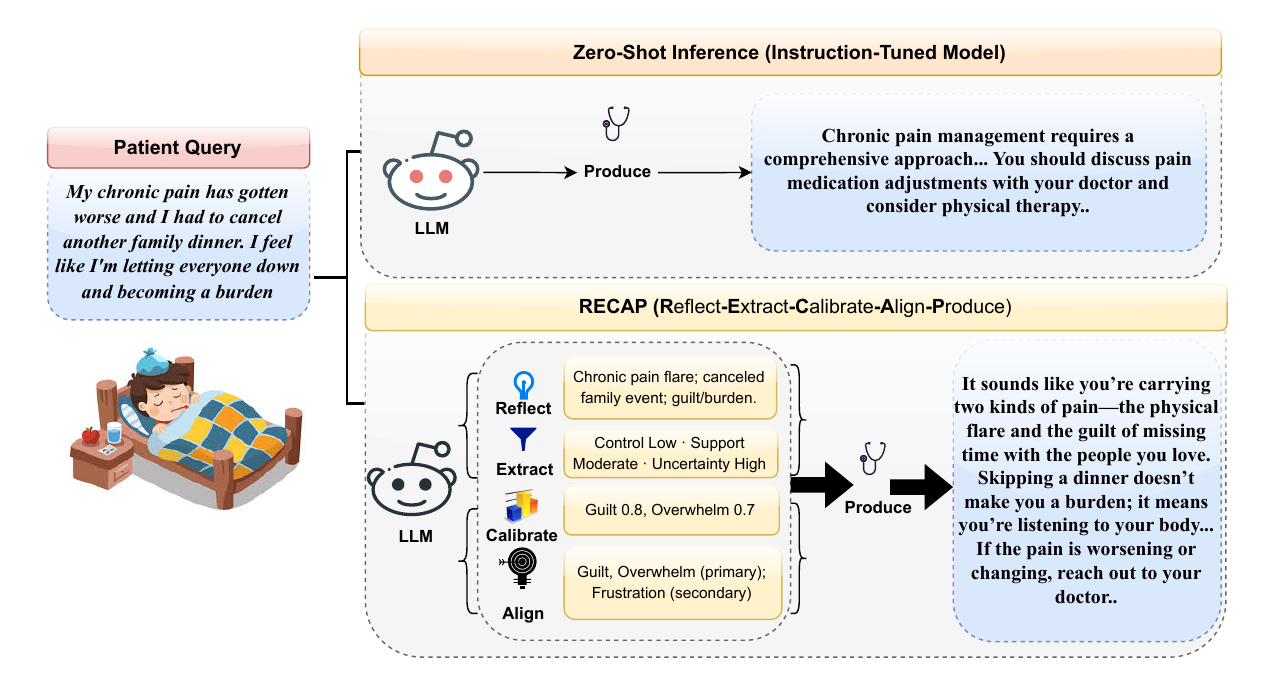}
  \caption{Patient input (left) is transformed into appraisal-theoretic intermediates with per-dimension Likert ratings (center), which condition generation to produce a targeted reply (right). Compared with a vanilla instruction-tuned model, RECAP explicitly acknowledges the patient's emotions and provides concrete next steps. \emph{(Scenario text is synthetic for illustration.)}}
  \label{fig:recap_teaser}
\end{figure*}

Cognitive appraisal theories suggest that humans evaluate events along key dimensions such as goal relevance, agency, and coping potential to determine emotional responses \citep{Ortony1988OCC,Lazarus1991EmotionAdaptation,NEURIPS2024_b0049c3f}. While LLMs do not experience emotions, recent work shows they can emulate aspects of this appraisal process when properly guided \citep{Croissant2024PLOSOne}. This presents an opportunity. By prompting models to produce appraisal-theoretic intermediates explicitly, we can both improve emotional alignment and examine how models reason about affect.

We introduce \method{} (\textbf{R}eflect--\textbf{E}xtract--\textbf{C}alibrate--\textbf{A}lign--\textbf{P}roduce), an inference-time pipeline that externalizes emotional reasoning into inspectable stages. \method{} guides models through structured reasoning, abstracting the situation, identifying psychological factors, generating candidate emotions, quantifying their likelihood via Likert scales, and producing an emotionally aligned response. Each stage yields interpretable artifacts such as situational abstractions, psychological factors, and emotion candidates with intensity ratings. These artifacts serve two purposes, providing clinicians with auditable checkpoints and enabling analysis of how models represent emotional content.

Our experiments reveal three insights about emotional reasoning in LLMs. First, explicit calibration improves alignment with human judgments, as \method{} reduces distance from human consensus by 20\% on SECEU. Second, structured reasoning disproportionately benefits smaller models (28\% improvement for 8B vs 5\% for 120B), suggesting explicit decomposition compensates for capabilities larger models have internalized. Third, examining intermediate outputs reveals systematic biases, with models consistently identifying control and uncertainty but underweighting relational factors like social support. Beyond these insights, \method{} improves practical performance. Oncology fellows rated \method{} responses significantly higher than baselines (3.41 vs 2.93--2.95, $p<.001$), with 76--88\% win rates across clinical scenarios.

\section{Background and Related Work}

\subsection{Empathy in Medical Dialogue}

LLMs are increasingly used for health information and triage. Multiple studies report that chatbot replies can be perceived as highly empathetic, sometimes surpassing physicians in perceived warmth while maintaining clinical accuracy \citep{Ayers2023JAMAIM,Chen2024JAMAOnc,Luo2024,Foyen2025}. However, systematic reviews find empathy to be brittle and prompt-sensitive, with overuse of generic phrases that undermine genuine connection \citep{Sorin2024JMIR,Armbruster2024,ye2025evaluatingmedicalllmslevels}. This raises a fundamental question about whether LLMs reason about emotional states or rely on surface-level stylistic mimicry. Answering this requires methods that externalize emotional reasoning into inspectable components.

\subsection{Approaches to Emotional Alignment}

Training-time methods including RLHF, DPO, and supervised fine-tuning on empathetic corpora can raise perceived empathy to clinician-rated levels \citep{Ouyang2022RLHF,Rafailov2023DPO,Rashkin2019EmpatheticDialogues,chen2023soulchat,Foyen2025}. However, these approaches embed empathetic behavior in model weights, making reasoning opaque and difficult to audit \citep{Beede2020CHI}. We cannot determine whether a fine-tuned model recognized the user's emotional state or simply learned empathetic-sounding language, a critical limitation for medical applications where transparency is not merely desirable but essential for safety, regulatory compliance, and clinical trust.

Inference-time approaches offer greater interpretability. Persona prompts like ``supportive clinician'' show mixed effectiveness \citep{Zhang2018PersonaChat,HuCollier2024PersonaEffect,Zheng2023PersonaPrompts,LuzdeAraujo2024PersonaBehavior}. Multi-step reasoning methods like Chain of Thought, Chain of Empathy, and Chain of Emotion improve emotional coherence through structured prompting \citep{Wei2022CoT,Lee2024CoE,Croissant2024PLOSOne}. Recent work demonstrates that decomposing inference into explicit steps improves both performance and auditability \citep{zhou2022learning,Wolfson2020BreakItDown,ThinkTuning,BOW,ToW,Tafjord2022Entailer,feng2024bird,Feng2023GenericTemporal,zhou2024conceptual,CCLearn}.

\method{} extends decomposition to emotional reasoning, grounding our pipeline in cognitive appraisal theory \citep{Ortony1988OCC,Lazarus1991EmotionAdaptation}. Unlike prior work, we expose per-dimension ratings that make emotional reasoning inspectable, enabling examination of whether LLMs apply consistent theories of emotion causation or exhibit systematic biases \citep{Amershi2019GuidelinesForHumanAI,Bucinca2021ToTrustOrToThink}.

\subsection{Transparency in Clinical AI}

Transparency is critical for AI systems making emotional assessments in healthcare. High technical accuracy alone is insufficient, and systems fail when reasoning is opaque or misaligned with clinical practices \citep{Beede2020CHI}. Systems with interpretable reasoning are more likely to be trusted and adopted \citep{Cai2019CSCW,Yang2019CHI,dineen2024unified,dineen2025qa,Dolk_2020}. \method{} addresses this by exposing intermediate artifacts, including situation abstractions, psychological factors, emotion candidates, and calibrated ratings, that clinicians can inspect, verify, or override.

\section{\method{}}

Understanding how LLMs process emotional content remains largely opaque. When a model produces an empathetic response, we cannot determine whether it recognized the user's emotional state, inferred relevant psychological context, or simply pattern-matched to training examples. \method{} addresses this by decomposing emotional reasoning into five explicit stages, each producing an inspectable intermediate artifact. This design serves two purposes. First, it improves response quality through structured reasoning. Second, it creates observation points that reveal how models represent and reason about affect.

We ground our decomposition in cognitive appraisal theory \cite{Ortony1988OCC,Lazarus1991EmotionAdaptation}, which posits that emotions arise from evaluations along dimensions such as goal relevance, agency, and coping potential. Our five-stage structure corresponds to theorized components of human emotional appraisal \cite{SchererMoors2019EmotionProcess}, namely situation encoding, relevance assessment, emotion categorization, intensity estimation, and response selection. Unlike chain-of-thought prompting, which produces unstructured traces that resist quantitative analysis, \method{}'s fixed schema enables aggregation across examples. We can compute statistics over factor distributions, emotion candidate frequencies, and calibration curves that reveal population-level patterns in model behavior. The pipeline requires no fine-tuning and executes in a single conversation turn. Figure~\ref{fig:recap_architecture} illustrates the complete pipeline with a representative clinical example.

To demonstrate each stage, we use a running example where a patient expresses the following.
\begin{quote}
\textit{``I was just diagnosed with cancer last month and the chemo is making me so sick. I'm scared this won't get any better. My sons keep asking me if I'm okay and I don't know what to say to them...''}
\end{quote}

\begin{figure*}[t!]
\centering

\definecolor{stageblue}{RGB}{65,125,190}
\definecolor{stageteal}{RGB}{55,165,155}
\definecolor{stageorange}{RGB}{225,130,55}
\definecolor{stagepurple}{RGB}{140,95,170}
\definecolor{stagegreen}{RGB}{80,160,85}
\definecolor{inputgray}{RGB}{85,85,95}

\begin{tikzpicture}
\node[rectangle, rounded corners=4pt, draw=inputgray!60, fill=inputgray!6, 
      line width=0.5pt,
      text width=0.83\textwidth, inner sep=6pt] (input) at (0,0) {
  \scriptsize\textcolor{inputgray}{\textbf{Patient Input:}}
  \textit{``I was just diagnosed with cancer last month and the chemo is making me so sick.
  I'm scared this won't get any better. My sons keep asking me if I'm okay and I don't know what to say to them...''}
};
\draw[-{Stealth[length=4pt, width=3pt]}, line width=1pt, draw=gray!50] (input.south) -- ++(0,-0.4);
\end{tikzpicture}

\vspace{0.1cm}

\small
\setlength{\tabcolsep}{3pt}
\begin{tabular}{@{}p{0.185\textwidth}@{\hspace{4pt}}p{0.185\textwidth}@{\hspace{4pt}}p{0.185\textwidth}@{\hspace{4pt}}p{0.185\textwidth}@{\hspace{4pt}}p{0.185\textwidth}@{}}

\cellcolor{stageblue}\textcolor{white}{\textbf{\footnotesize 1. REFLECT}} &
\cellcolor{stageteal}\textcolor{white}{\textbf{\footnotesize 2. EXTRACT}} &
\cellcolor{stageorange}\textcolor{white}{\textbf{\footnotesize 3. CALIBRATE}} &
\cellcolor{stagepurple}\textcolor{white}{\textbf{\footnotesize 4. ALIGN}} &
\cellcolor{stagegreen}\textcolor{white}{\textbf{\footnotesize 5. PRODUCE}} \\

\cellcolor{stageblue!6}
\scriptsize\textcolor{stageblue!70!black}{\textit{Situation Abstraction}}\newline\vspace{2pt}
``The patient has been diagnosed with a serious illness and is undergoing difficult treatment, causing them to feel overwhelmed and have anxiety about the future.''
&
\cellcolor{stageteal!6}
\scriptsize\textcolor{stageteal!70!black}{\textit{Factor Induction}}\newline\vspace{2pt}
\textbf{Control:} Low\newline{\tiny\textit{(little control over illness)}}\newline
\textbf{Support:} Moderate\newline{\tiny\textit{(family present)}}\newline
\textbf{Uncertainty:} High\newline{\tiny\textit{(unsure of outcomes)}}
&
\cellcolor{stageorange!6}
\scriptsize\textcolor{stageorange!70!black}{\textit{Candidate Emotions}}\newline\vspace{2pt}
• Anxiety\newline
• Fear\newline
• Overwhelm/Stress\newline
• Sadness\newline
• Hope\newline
• Guilt\newline
• Helplessness
&
\cellcolor{stagepurple!6}
\scriptsize\textcolor{stagepurple!70!black}{\textit{Likert Rating}}\newline\vspace{2pt}
Anxiety: \textbf{0.95}\newline
Fear: \textbf{0.95}\newline
Overwhelm: \textbf{0.95}\newline
Sadness: \textbf{0.50}\newline
Hope: \textbf{0.25}\newline
Guilt: \textbf{0.75}\newline
Helplessness: \textbf{0.75}
&
\cellcolor{stagegreen!6}
\scriptsize\textcolor{stagegreen!70!black}{\textit{Aligned Response}}\newline\vspace{2pt}
\textit{``I'm really sorry you're feeling so overwhelmed and anxious. That's completely understandable---facing serious illness can be frightening...''}
\\
\end{tabular}

\caption{\textbf{RECAP Pipeline for Emotional Alignment.} Model-agnostic inference-time prompting that externalizes emotional reasoning into interpretable stages: (1) abstraction, (2) factor identification, (3) emotion enumeration, (4) Likert assessment, and (5) aligned response generation.}
\label{fig:recap_architecture}
\end{figure*}

\subsection{Situation Understanding (Stages 1--2)}

\paragraph{Stage 1, Situation Abstraction.} The pipeline first distills raw input into a concise description of the core situation, stripping lengthy narratives to capture essential circumstances \cite{Sweller2023CLTExpansion,feng2024bird}. The abstraction reveals what the model considers emotionally salient. Abstractions that omit key triggers (e.g., family concerns in Figure~\ref{fig:recap_architecture}) signal gaps in emotional scene understanding.

\begin{quote}
\textit{Example: ``The patient has been diagnosed with a serious illness and is undergoing difficult treatment, which makes them feel overwhelmed and anxious about the future.''}
\end{quote}

\paragraph{Stage 2, Latent Factor Induction.} The model identifies three psychological factors that shape emotional responses, assigning values and justifications. These constructs directly map to core appraisal theory dimensions, specifically \textit{primary appraisal} (assessing goal relevance and congruence) and \textit{secondary appraisal} (evaluating agency, control, and coping potential) \cite{SchererMoors2019EmotionProcess, Lazarus1991EmotionAdaptation}. Factor induction exposes the model's implicit theory of emotion causation. The factors selected, and those not selected, reveal which dimensions the model treats as emotionally relevant, enabling comparison with human appraisal patterns and identifying systematic biases that propagate downstream. For the running example, the induced factors are as follows.
\begin{itemize}[noitemsep]
\item \textbf{Control (Secondary Appraisal)} Low, as the patient has little control over illness or treatment.
\item \textbf{Support (Coping Potential)} Moderate, as family is present but the patient feels isolated.
\item \textbf{Outcome Uncertainty (Primary Appraisal)} High, as the patient is unsure about treatment success.
\end{itemize}

By limiting to three factors, we ensure focus on the most salient dimensions while maintaining adaptability across situations. Open-vocabulary induction avoids dependence on any fixed ontology, allowing factors to vary naturally (e.g., ``financial strain'' for one patient, ``parental responsibility'' for another) while remaining recognizable to clinicians familiar with appraisal-based reasoning \cite{Amershi2019GuidelinesForHumanAI}.

\subsection{Emotion Modeling (Stages 3--4)}

\paragraph{Stage 3, Candidate Emotion Extraction.} With situation and factors established, \method{} generates 5--7 plausible candidate emotions rather than committing to a single label, reflecting that real-world interactions evoke mixed affect \cite{HepburnPotter2023Emotionography}. The candidate set reveals the model's emotion ontology. Systematic patterns such as consistently generating ``anxiety'' for medical contexts regardless of patient agency indicate stereotyped associations rather than context-sensitive reasoning. For the running example, candidates include anxiety, fear, overwhelm, sadness, hope, guilt, and helplessness.

\paragraph{Stage 4, Likert-Based Calibration.} Each candidate is rated on a five-point scale (very-unlikely to very-likely, mapped to probabilities 0.05--0.95), capturing partial belief and mixed affect rather than forcing exclusivity. This addresses the failure mode where LLMs guess rather than express uncertainty \cite{Kalai2025WhyLM,kadavath2022know}. The resulting profile is a compact, clinician-reviewable vector. Well-calibrated models should assign high ratings to human-endorsed emotions and mid-range ratings for ambiguous situations. Miscalibration patterns reveal reasoning deficits that benchmark accuracy alone cannot detect.

\subsection{Response Generation (Stage 5)}

Guided by the emotion-likelihood profile, the model acknowledges the highest-probability feelings, validates them, and adjusts tone before offering guidance, following clinical frameworks that prioritize emotions before information \cite{FRANCIS2023107574}. Conditioning on explicit profiles also enables counterfactual probing, allowing us to test whether varying Likert input predictably changes response tone. This causal testing of emotion-to-language mappings is impossible with end-to-end generation. The modular design additionally supports clinical override, because emotional guidance can be adjusted based on external input.

\begin{quote}
\textit{``I'm really sorry you're feeling so overwhelmed and anxious about your diagnosis and treatment. That's completely understandable---facing serious illness can be frightening and stressful. You're not alone in this...''}
\end{quote}
\section{Experiments}
\label{sec:experiments}

\subsection{Datasets}

We evaluate \method{} on benchmarks spanning single-turn reasoning and multi-turn dialogue.

\paragraph{Single-Turn Datasets}

\begin{itemize}[noitemsep]
\item \textbf{EmoBench} \citep{Sabour2024EmoBench}: Complex scenarios requiring inference of latent affect and causal emotional reasoning. Unlike simple emotion classification, EmoBench emphasizes reasoning about multi-party emotional dynamics and implicit cues.
\item \textbf{SECEU} \citep{Wang2023EmotionalIntelligenceLLMs}: Emotion intensity distribution task where models allocate relative intensity across candidate emotions. Reports EQ scores (higher is better) and raw distance from human consensus (lower is better).
\end{itemize}

\paragraph{Multi-Turn Datasets}

\begin{itemize}[noitemsep]
\item \textbf{EQ-Bench3} \citep{eqbench}: Dialogue-centric emotional inference with Elo-style ratings, testing emotional reasoning in conversational contexts with social dynamics.
\item \textbf{HealthBench} \citep{arora2025healthbench}: 5,000 multi-turn healthcare conversations scored against physician-written rubrics spanning accuracy, completeness, context awareness, communication quality, and instruction following.
\end{itemize}

\subsection{Baselines}

We compare \method{} against three prompting strategies.

\begin{enumerate}[noitemsep]
\item \textbf{Zero-shot}: Task description and input only, without emotion-specific guidance. This tests the model's inherent emotional understanding acquired during pretraining and instruction tuning.
\item \textbf{Emotion Priming}: Augments the zero-shot prompt with explicit instructions to analyze emotional states and cues, for example ``Deeply analyze the provided story, focusing on the main character's situation, actions, and any explicit or implicit emotional cues.'' This tests whether directing attention to emotional analysis improves performance without structured reasoning.
\item \textbf{Supportive Clinician Persona}: Role-based conditioning as ``an empathetic, supportive clinician'' tasked with evaluating the emotional state of the person in the scenario. Unlike \method{}, this relies on implicit conditioning rather than explicit emotional inference.
\end{enumerate}

We exclude fine-tuned models (e.g., SoulChat \citep{chen2023soulchat}) to isolate \method{}'s inference-time contribution under zero-training constraints.

\subsection{Models and Configuration}

Experiments use Llama-3.1-8B-Instruct \citep{llama31}, OpenAI-OSS-20B, and OpenAI-OSS-120B \citep{gptoss}. We focus on open-source models for three reasons. First, cost efficiency, as \method{}'s multi-step reasoning makes API costs prohibitive. Second, reproducibility. Third, competitive performance, since OSS-120B achieves near-parity with o4-mini on reasoning benchmarks. All experiments use temperature 0.6, max tokens 2048, with three trials per condition.

\subsection{Clinical Evaluation Setup}

We conducted blind evaluations with two board-certified oncology research fellows, comparing \method{} against all baselines for emotional support quality.

\begin{table*}[t!]
\centering
\small

\begin{subtable}[t]{\textwidth}
\centering
\caption{Aggregate emotional-intelligence benchmarks}
\setlength{\tabcolsep}{3pt}
\begin{tabular}{@{}l|ccc|ccc|ccc@{}}
\toprule
& \multicolumn{3}{c|}{\textbf{EmoBench} (Acc.$\uparrow$)} & \multicolumn{3}{c|}{\textbf{SECEU} (EQ$\uparrow$)} & \multicolumn{3}{c}{\textbf{EQ-Bench3} (Elo$\uparrow$)} \\
\textbf{Method} & L3-8B & O-20B & O-120B & L3-8B & O-20B & O-120B & L3-8B & O-20B & O-120B \\
\midrule
Zero-Shot        & $.35_{\pm.1}$ & $.61_{\pm.1}$ & $.78_{\pm.1}$ & $58.3_{\pm3.9}$ & $96.4_{\pm4.1}$ & $110.5_{\pm2.2}$ & $1032_{\pm1}$ & $1281_{\pm2}$ & $1474_{\pm1}$ \\
Emotion Priming  & $\underline{.42_{\pm.2}}$ & $\mathbf{.65_{\pm.1}}$ & $\mathbf{.83_{\pm.1}}$ & $\underline{61.7_{\pm3.9}}$ & $97.4_{\pm3.5}$ & $\underline{111.1_{\pm2.2}}$ & $\underline{1033_{\pm2}}$ & $\underline{1284_{\pm3}}$ & $1475_{\pm2}$ \\
Persona          & $.39_{\pm.1}$ & $.64_{\pm.1}$ & $\underline{.81_{\pm.1}}$ & $58.9_{\pm5.8}$ & $\underline{100.4_{\pm2.2}}$ & $109.8_{\pm2.5}$ & $1032_{\pm1}$ & $1283_{\pm3}$ & $\underline{1476_{\pm2}}$ \\
\rowcolor{blue!6}
\textbf{RECAP}   & $\mathbf{.45_{\pm.2}}$ & $\underline{.65_{\pm.2}}$ & $\mathbf{.83_{\pm.1}}$ & $\mathbf{71.3_{\pm4.1}}$ & $\mathbf{109.2_{\pm3.6}}$ & $\mathbf{115.2_{\pm3.8}}$ & $\mathbf{1101_{\pm3}}$ & $\mathbf{1308_{\pm3}}$ & $\mathbf{1484_{\pm3}}$ \\
\bottomrule
\end{tabular}
\end{subtable}

\vspace{0.8em}

\begin{subtable}[t]{0.85\textwidth}
\centering
\caption{SECEU breakdown: EQ score vs. raw distance from consensus}
\setlength{\tabcolsep}{4pt}
\begin{tabular}{@{}l|cc|cc|cc@{}}
\toprule
& \multicolumn{2}{c|}{\textbf{L3-8B}} & \multicolumn{2}{c|}{\textbf{O-20B}} & \multicolumn{2}{c}{\textbf{O-120B}} \\
\textbf{Method} & EQ$\uparrow$ & Dist.$\downarrow$ & EQ$\uparrow$ & Dist.$\downarrow$ & EQ$\uparrow$ & Dist.$\downarrow$ \\
\midrule
Zero-Shot        & $58.3_{\pm3.9}$ & $5.1_{\pm0.7}$ & $96.4_{\pm4.1}$ & $3.1_{\pm0.3}$ & $110.5_{\pm2.2}$ & $\mathbf{2.2_{\pm0.3}}$ \\
Emotion Priming  & $\underline{61.7_{\pm3.9}}$ & $4.8_{\pm0.5}$ & $97.4_{\pm3.5}$ & $2.7_{\pm0.4}$ & $\underline{111.1_{\pm2.2}}$ & $\underline{2.4_{\pm0.3}}$ \\
Persona          & $58.9_{\pm5.8}$ & $\underline{4.8_{\pm0.7}}$ & $\underline{100.4_{\pm2.2}}$ & $\underline{2.8_{\pm0.6}}$ & $109.8_{\pm2.5}$ & $2.4_{\pm0.4}$ \\
\rowcolor{blue!6}
\textbf{RECAP}   & $\mathbf{71.3_{\pm4.1}}$ & $\mathbf{4.1_{\pm0.8}}$ & $\mathbf{109.2_{\pm3.6}}$ & $\mathbf{2.5_{\pm0.5}}$ & $\mathbf{115.2_{\pm3.8}}$ & $2.4_{\pm0.6}$ \\
\bottomrule
\end{tabular}
\end{subtable}

\caption{Emotional intelligence results. Models: L3-8B = Llama-3.1-8B-Instruct, O-20B = GPT-OSS-20B, O-120B = GPT-OSS-120B. (a) Aggregate scores on EmoBench, SECEU, and EQ-Bench3. (b) SECEU decomposition showing EQ score and distance from human consensus (lower = better). \textbf{Bold} = best; \underline{underline} = second best. All runs: temp=0.6, max\_tokens=2048, $n$=3 trials. Values are mean$_{\pm\text{sem}}$.}
\label{tab:emotional-intelligence-combined}
\end{table*}

\paragraph{Data Curation.} We synthesized evaluation scenarios based on authentic clinical interactions. Source material consisted of anonymized patient charts and questions from chemotherapy consultations, compiled by oncology fellows through medical records review, chemotherapy educational documentation, and safety resources. Using this corpus, GPT-5 generated structurally similar scenarios preserving clinical complexity, and all underwent clinician review and approval.

\begin{figure}[t]
    \centering
    \begin{tcolorbox}[
        enhanced,
        title={\textbf{(a) Single-Turn Scenario}},
        fonttitle=\scriptsize,
        coltitle=white,
        colbacktitle=blue!55!black,
        colback=blue!3,
        colframe=blue!40!black,
        boxrule=0.5pt,
        arc=2pt,
        left=4pt, right=4pt, top=3pt, bottom=3pt,
    ]
    \scriptsize
    \textcolor{blue!60!black}{\textbf{Dx}} De novo HER2+ metastatic breast cancer with gastric/cecal involvement\par\vspace{1.5pt}
    \textcolor{blue!60!black}{\textbf{Tx}} Docetaxel + trastuzumab + pertuzumab q3w, limit docetaxel to 4--6 cycles, then maintenance\par\vspace{1.5pt}
    \textcolor{blue!60!black}{\textbf{Pt}} 60-y/o grandmother, lost 30\,lb in 3\,mo. Fears losing her identity to her grandchildren.\par\vspace{1.5pt}
    \textcolor{blue!60!black}{\textbf{Q}} \emph{``Everyone keeps saying this plan is palliative. How do I balance hope with honesty so my family doesn't think I'm giving up?''}
    \end{tcolorbox}

    \vspace{3pt}

    \begin{tcolorbox}[
        enhanced,
        title={\textbf{(b) Multi-Turn Scenario}},
        fonttitle=\scriptsize,
        coltitle=white,
        colbacktitle=teal!60!black,
        colback=teal!3,
        colframe=teal!45!black,
        boxrule=0.5pt,
        arc=2pt,
        left=4pt, right=4pt, top=3pt, bottom=3pt,
    ]
    \scriptsize
    \textcolor{teal!65!black}{\textbf{Dx}} Stage IIA classical Hodgkin lymphoma\par\vspace{1.5pt}
    \textcolor{teal!65!black}{\textbf{Tx}} ABVD chemotherapy q2w for 2--4 cycles with response-adapted PET imaging\par\vspace{1.5pt}
    \textcolor{teal!65!black}{\textbf{Pt}} 20-y/o first-gen college student. Believes \emph{``I cannot fail. My parents sacrificed everything.''}\par\vspace{1.5pt}
    \textcolor{teal!65!black}{\textbf{Goals}} Refuse semester off, seek strategies to study through cognitive effects\par\vspace{1.5pt}
    \textcolor{teal!65!black}{\textbf{Opening}} \emph{``I finally made the Dean's List. I can't take a semester off. My parents worked too hard for me to be here.''}
    \end{tcolorbox}

    \caption{Representative synthetic patient scenarios. (a)~Single-turn evaluation assesses individual response quality. (b)~Multi-turn evaluation tracks conversational dynamics across 3 turns.}
    \label{fig:scenarios}
    \label{fig:scenario-single}
    \label{fig:scenario-multi}
\end{figure}

Figure~\ref{fig:scenarios} shows representative examples. For single-turn evaluation, we constructed 25 scenarios comprising a diagnosis, a treatment plan, a clinical narrative, and a patient question. For multi-turn evaluation, 10 scenarios included synthetic patient personas with demographics, personality traits, and hidden conversational goals guiding realistic behavior across three turns. GPT-5 simulated patient responses, with persona and goals enabling dynamic adaptation. Both datasets were stratified across cancer types and emotional states. Our design aligns with work on LLM-based simulated patients \citep{wang2024patientpsiusinglargelanguage,lee2025adaptivevpframeworkllmbasedvirtual,bodonhelyi2025modelingchallengingpatientinteractions}, embedding persona-level beliefs that support coherent emotional trajectories.

\paragraph{Annotation Process.} Clinical annotators had no prior knowledge of \method{}'s design. Responses were presented in randomized, blinded order unique to each annotator. Each response was rated independently on Likert scales (1--5) with optional qualitative feedback. We conducted two evaluation tasks.

\begin{itemize}[noitemsep]
\item \textbf{Single-turn}: 100 responses across 25 scenarios, four methods each, five evaluation dimensions (Completeness, Emotional Recognition, Supportiveness, Trustworthiness, Tone Appropriateness).
\item \textbf{Multi-turn}: 20 conversations (10 scenarios $\times$ 2 methods), three turns each, four dimensions (Personalization, Context Awareness, Tone, Coherence). Compared \method{} against Zero-Shot only, as single-turn findings showed all three baselines performed equivalently (2.93--2.95).
\end{itemize}

Both annotators independently rated all 100 single-turn responses. Inter-annotator agreement yielded 78.2\% within-one-point agreement across five dimensions (quadratic weighted $\kappa$=0.09, Spearman $\rho$=0.20, $p<.001$). The modest $\kappa$ reflects differing annotator calibration (mean ratings of 3.50 vs.\ 2.61) rather than disagreement on relative ordering. We report averaged scores throughout.

\section{Results}
\label{sec:results}

\subsection{Emotion Understanding Benchmarks}

Table~\ref{tab:emotional-intelligence-combined} presents results across three emotional intelligence benchmarks and three model scales. \method{} demonstrates consistent improvements, particularly pronounced on tasks that require a nuanced understanding of emotional states.

\paragraph{EmoBench.}

For Llama-3.1-8B-Instruct, \method{} achieves 0.45 accuracy compared to 0.35 for zero-shot (28\% relative improvement). Emotion Priming yields intermediate performance (0.423), indicating that structured reasoning offers additional benefits beyond emotional activation. At OSS-120B, \method{} and Emotion Priming both reach 0.825 versus the zero-shot baseline (0.783). The Persona baseline shows mixed results, occasionally underperforming zero-shot, suggesting role-based prompting without structured reasoning may introduce biases.

\paragraph{SECEU.}

\method{} achieves EQ scores of 71.29, 109.17, and 115.19 across model scales, consistently outperforming all baselines (22\% improvement over zero-shot at 8B). \method{} also achieves the lowest distance from human consensus (4.08 vs 5.08, a 20\% improvement), indicating that explicit Likert calibration produces emotion estimates more aligned with human judgments. At OSS-120B scale, zero-shot achieves lower raw distance (2.22) while \method{} still produces the highest EQ score (115.19). This reflects the EQ scoring mechanism, which rewards not just accuracy but appropriate confidence calibration, suggesting \method{} helps models express uncertainty more appropriately.

\paragraph{EQ-Bench3.}

\method{} demonstrates substantial improvements across scales, including a 70-point Elo gain for Llama-3.1-8B (1101.03 vs 1031.92), 28 points for OSS-20B (1308.24 vs 1280.57), and 10 points for OSS-120B (1483.7 vs 1473.7). These consistent gains suggest structured emotional reasoning provides value even for highly capable models, though diminishing returns at scale indicate larger models have stronger implicit emotional reasoning.

\subsection{Medical Communication Quality}

Table~\ref{tab:hb-combined} presents HealthBench overall scores (per-dimension breakdown in Table~\ref{tab:hb-full}, Appendix). \method{} improves overall quality at every scale, with gains of 8\% for Llama-3.1-8B, 11\% for OSS-20B, and 18\% for OSS-120B. The largest gains occur in Completeness (23\% for OSS-20B) and Communication Quality (27\% for OSS-20B), suggesting emotional reasoning helps models address both medical and emotional dimensions of patient concerns.

However, \method{} exhibits systematic trade-offs. Accuracy and Instruction Following decrease across all scales, reflecting prioritization of comprehensive, emotionally tuned responses over literal precision. Increased standard deviations (e.g., Llama-3.1-8B from 0.0067 to 0.0235) indicate more variable, scenario-specific responses rather than formulaic outputs. These accuracy differences do not reflect knowledge limitations, as \method{} uses the same underlying models, but rather a shift toward emotional appropriateness in response generation. This surfaces a style-versus-precision tension common to empathy-forward systems, mitigable through verification techniques such as self-refinement.

\begin{table}[t]
\centering
\small
\begin{tabular}{@{}lccc@{}}
\toprule
\textbf{Method} & \textbf{L3-8B} & \textbf{O-20B} & \textbf{O-120B} \\
\midrule
Zero-Shot       & $0.18_{\pm0.01}$ & $0.37_{\pm0.02}$ & $0.46_{\pm0.01}$ \\
Emotion Priming & $0.18_{\pm0.01}$ & $0.38_{\pm0.03}$ & $0.49_{\pm0.03}$ \\
Persona         & $0.19_{\pm0.02}$ & $0.40_{\pm0.04}$ & $0.52_{\pm0.04}$ \\
\rowcolor{blue!6}
\textbf{RECAP}  & $\mathbf{0.19_{\pm0.02}}$ & $\mathbf{0.41_{\pm0.04}}$ & $\mathbf{0.54_{\pm0.05}}$ \\
\bottomrule
\end{tabular}
\caption{HealthBench overall scores across model scales. L3-8B = Llama-3.1-8B-Instruct, O-20B = GPT-OSS-20B, O-120B = GPT-OSS-120B. Per-dimension breakdown in Table~\ref{tab:hb-full} (Appendix). Values are mean$_{\pm\text{sem}}$; $n$=3 trials.}
\label{tab:hb-combined}
\end{table}

\begin{figure*}[t]
    \centering
    \pgfplotsset{
        human eval/.style={
            ybar,
            width=\linewidth,
            height=4.5cm,
            ylabel={Rating (1--5)},
            ymin=2.3, ymax=4.7,
            ytick={2.5, 3.0, 3.5, 4.0, 4.5},
            ylabel style={font=\small},
            xlabel style={font=\small},
            x tick label style={rotate=25, anchor=east, font=\footnotesize},
            y tick label style={font=\footnotesize},
            xtick=data,
            axis line style={gray!50},
            major tick style={gray!50},
            ymajorgrids=true,
            grid style={gray!15, dashed},
            axis on top,
        },
    }
    
    \begin{subfigure}[b]{0.54\textwidth}
        \centering
        \begin{tikzpicture}
            \begin{axis}[
                human eval,
                bar width=6pt,
                symbolic x coords={Completeness, Emot. Recog., Supportiveness, Trustworth., Tone Approp.},
                enlarge x limits=0.1,
            ]
            \addplot[fill=blue!60, draw=blue!70!black, error bars/.cd, y dir=plus, y explicit, error bar style={black!50, thick}] coordinates {
                (Completeness, 3.50) +- (0, 0.08)
                (Emot. Recog., 3.18) +- (0, 0.09)
                (Supportiveness, 3.42) +- (0, 0.10)
                (Trustworth., 3.28) +- (0, 0.09)
                (Tone Approp., 3.66) +- (0, 0.09)
            };
            \addplot[fill=gray!45, draw=gray!60!black, error bars/.cd, y dir=plus, y explicit, error bar style={black!50, thick}] coordinates {
                (Completeness, 3.18) +- (0, 0.12)
                (Emot. Recog., 2.58) +- (0, 0.11)
                (Supportiveness, 2.78) +- (0, 0.12)
                (Trustworth., 3.00) +- (0, 0.11)
                (Tone Approp., 3.20) +- (0, 0.11)
            };
            \addplot[fill=orange!55, draw=orange!70!black, error bars/.cd, y dir=plus, y explicit, error bar style={black!50, thick}] coordinates {
                (Completeness, 3.18) +- (0, 0.11)
                (Emot. Recog., 2.64) +- (0, 0.09)
                (Supportiveness, 2.74) +- (0, 0.10)
                (Trustworth., 2.86) +- (0, 0.09)
                (Tone Approp., 3.24) +- (0, 0.10)
            };
            \addplot[fill=teal!55, draw=teal!70!black, error bars/.cd, y dir=plus, y explicit, error bar style={black!50, thick}] coordinates {
                (Completeness, 3.28) +- (0, 0.11)
                (Emot. Recog., 2.54) +- (0, 0.10)
                (Supportiveness, 2.72) +- (0, 0.12)
                (Trustworth., 3.02) +- (0, 0.12)
                (Tone Approp., 3.12) +- (0, 0.11)
            };
            \end{axis}
        \end{tikzpicture}
        \caption{Single-turn ratings (n=25 per method)}
    \end{subfigure}
    \hfill
    \begin{subfigure}[b]{0.42\textwidth}
        \centering
        \begin{tikzpicture}
            \begin{axis}[
                human eval,
                bar width=8pt,
                symbolic x coords={Personal., Context, Tone, Coherence},
                enlarge x limits=0.18,
            ]
            \addplot[fill=blue!60, draw=blue!70!black, error bars/.cd, y dir=plus, y explicit, error bar style={black!50, thick}] coordinates {
                (Personal., 3.95) +- (0, 0.05)
                (Context, 3.80) +- (0, 0.14)
                (Tone, 3.55) +- (0, 0.15)
                (Coherence, 3.80) +- (0, 0.12)
            };
            \addplot[fill=gray!45, draw=gray!60!black, error bars/.cd, y dir=plus, y explicit, error bar style={black!50, thick}] coordinates {
                (Personal., 3.80) +- (0, 0.09)
                (Context, 3.70) +- (0, 0.15)
                (Tone, 3.60) +- (0, 0.13)
                (Coherence, 3.75) +- (0, 0.14)
            };
            \end{axis}
        \end{tikzpicture}
        \caption{Multi-turn ratings (n=10 per method)}
    \end{subfigure}
    
    \vspace{0.4em}
    \ref*{legend:human-eval-main}

    \begin{tikzpicture}[baseline]
        \begin{axis}[
            hide axis,
            xmin=0, xmax=1, ymin=0, ymax=1,
            legend style={
                draw=none,
                legend columns=5,
                font=\footnotesize,
                /tikz/every even column/.append style={column sep=5pt},
            },
            legend to name=legend:human-eval-main,
        ]
        \addlegendimage{fill=blue!60, draw=blue!70!black, area legend}
        \addlegendentry{RECAP}
        \addlegendimage{fill=gray!45, draw=gray!60!black, area legend}
        \addlegendentry{Zero-Shot}
        \addlegendimage{fill=orange!55, draw=orange!70!black, area legend}
        \addlegendentry{E-PRIME}
        \addlegendimage{fill=teal!55, draw=teal!70!black, area legend}
        \addlegendentry{Persona}
        \end{axis}
    \end{tikzpicture}

    \caption{Human evaluation results. (a,b) Mean ratings with standard error bars (1--5 scale). Scenario-level win rates are reported in Figure~\ref{fig:human-eval-winrates} (Appendix).}
    \label{fig:human-eval}
\end{figure*}

\subsection{Clinical Evaluation Results}

\paragraph{Single-Turn Evaluation}

\method{} achieved an overall score of 3.41 ($\pm$0.32), significantly outperforming Zero-Shot (2.95, $p<.001$, $d$=0.98), Persona (2.94, $p<.001$, $d$=1.02), and Emotion Priming (2.93, $p<.001$, $d$=1.08). As shown in Figure~\ref{fig:human-eval}a, \method{} consistently outperforms all baselines across evaluation dimensions. 

The largest improvements emerged in empathy-related dimensions, specifically Supportiveness (+0.67, $d$=1.37) and Emotional Recognition (+0.59, $d$=1.34), both representing large effect sizes. Factual Completeness showed more modest improvement (+0.29, $d$=0.85), suggesting \method{}'s primary advantage lies in emotional rather than informational quality. All conditions scored lowest on Emotional Recognition (range 2.54--3.18), indicating this dimension remains challenging for LLMs.

Scenario-level win rates (Figure~\ref{fig:human-eval-winrates}, Appendix) show \method{} wins 88\% against Persona, 84\% against Emotion Priming, and 76\% against Zero-Shot. \method{} achieved the highest rating across all three baselines in 17 of 25 scenarios (68\%), with 46\% of responses rated 4--5 compared to 22--29\% for baselines.

\paragraph{Multi-Turn Evaluation}

\method{} maintained its advantage with an overall score of 3.77 ($\pm$0.29) versus Zero-Shot's 3.71 ($\pm$0.29). Figure~\ref{fig:human-eval}b shows \method{} outperforms Zero-Shot across all four dimensions, with the strongest gains on Personalization (3.95 vs 3.80, $d$=0.70) and greater consistency (SD=0.16 vs 0.26). The narrower gap compared to single-turn settings indicates that extended context provides additional grounding, though \method{} continues yielding benefits in adapting to individual patient needs. Scenario-level comparisons showed \method{} winning 5 of 10 scenarios (50\%), with 1 tie and 4 Zero-Shot wins, and achieving slightly higher-quality ratings (76\% rated 4--5 vs 72\% for Zero-Shot).

\paragraph{Qualitative Analysis}

VADER sentiment analysis \citep{hutto2014vader} of annotator comments showed \method{} received significantly more positive sentiment, with 68\% positive versus 28--56\% for baselines and mean compound scores of +0.142 versus -0.259 to -0.006 (Cohen's $d$=0.56 vs.\ pooled baselines). \method{} had zero scenarios rated below 2.5, compared to 1--3 for baselines, and achieved high ratings ($\geq$3.5) in 44\% of scenarios versus 8--12\% for baselines (Figure~\ref{fig:qual-bar}, Appendix). Annotators characterized \method{} responses as ``emotionally cognizant, supportive and complete,'' while criticizing baselines as ``robotic'' and ``medically correct but emotionally inadequate.''

\paragraph{LLM-as-a-Judge Evaluation}
\begin{figure}[H]
    \centering
    \begin{tikzpicture}
        \begin{axis}[
            width=0.95\linewidth,
            height=4.8cm,
            xlabel={Rating (1--5)},
            xmin=2.5, xmax=5.1,
            xtick={2.5, 3.0, 3.5, 4.0, 4.5, 5.0},
            symbolic y coords={Tone,Trust,Support,Emot. Rec.,Complete},
            ytick=data,
            y tick label style={font=\footnotesize},
            x tick label style={font=\footnotesize},
            xlabel style={font=\small},
            enlarge y limits=0.18,
            clip=false,
            axis line style={gray!50},
            major tick style={gray!50},
            xmajorgrids=true,
            grid style={gray!12, densely dotted},
            legend style={
                at={(0.5,-0.26)},
                anchor=north,
                legend columns=6,
                font=\tiny,
                column sep=2pt,
                draw=none,
                fill=none,
                overlay,
            },
        ]
        \addplot[gray!25, line width=3pt, forget plot, opacity=0.6] coordinates {(3.31,Tone) (4.71,Tone)};
        \addplot[gray!25, line width=3pt, forget plot, opacity=0.6] coordinates {(3.04,Trust) (4.89,Trust)};
        \addplot[gray!25, line width=3pt, forget plot, opacity=0.6] coordinates {(2.92,Support) (4.39,Support)};
        \addplot[gray!25, line width=3pt, forget plot, opacity=0.6] coordinates {(2.74,Emot. Rec.) (4.05,Emot. Rec.)};
        \addplot[gray!25, line width=3pt, forget plot, opacity=0.6] coordinates {(3.29,Complete) (4.85,Complete)};

        \addplot[only marks, mark=*, mark size=4pt, black, mark options={fill=white, line width=1.2pt}] coordinates {
            (3.31,Tone) (3.04,Trust) (2.92,Support) (2.74,Emot. Rec.) (3.29,Complete)
        };
        \addplot[only marks, mark=*, mark size=3pt, orange!85!red] coordinates {
            (4.71,Tone) (4.89,Trust) (4.39,Support) (4.05,Emot. Rec.) (4.85,Complete)
        };
        \addplot[only marks, mark=*, mark size=3pt, red!65!orange] coordinates {
            (4.19,Tone) (3.82,Trust) (3.81,Support) (3.58,Emot. Rec.) (3.85,Complete)
        };
        \addplot[only marks, mark=*, mark size=3pt, violet!70] coordinates {
            (4.33,Tone) (4.04,Trust) (4.13,Support) (3.95,Emot. Rec.) (4.02,Complete)
        };
        \addplot[only marks, mark=*, mark size=3pt, blue!65] coordinates {
            (3.38,Tone) (4.28,Trust) (3.70,Support) (3.12,Emot. Rec.) (4.36,Complete)
        };
        \addplot[only marks, mark=*, mark size=3pt, teal!70] coordinates {
            (3.80,Tone) (4.45,Trust) (3.66,Support) (3.45,Emot. Rec.) (4.12,Complete)
        };
        \legend{Human, 4o-mini, 5-mini, 5-nano, Sonnet, Haiku}
        \end{axis}
    \end{tikzpicture}
    \caption{LLM-as-Judge vs.\ human ratings. Scores represent means across all 200 single-turn responses (25 scenarios $\times$ 4 methods). Scenario quality distribution in Figure~\ref{fig:qual-bar}.}
    \label{fig:llm-judge}
\end{figure}

Five frontier LLMs (GPT-4o-mini, GPT-5-mini, GPT-5-nano, Claude Sonnet 4.5, Claude Haiku 4.5) rated all 100 single-turn responses. As shown in Figure~\ref{fig:llm-judge}, all models showed weak-to-moderate correlation with human judgments ($r$=0.09--0.26), with Emotional Recognition strongest ($r$=0.27--0.44) while Completeness and Trustworthiness showed near-zero or negative correlations ($r$ = -0.11 to 0.05). 

All models exhibited systematic positive bias, with GPT-4o-mini inflating most (+1.3--1.9 points) and GPT-5-mini showing the smallest bias (+0.3--0.9 points). Notably, lower bias did not correlate with better alignment: Claude Sonnet 4.5 showed moderate bias but the lowest correlation ($r$=0.09), suggesting its ratings captured a different signal than human judgments rather than simply being better calibrated.

\subsection{Cross-Benchmark Analysis}

Several consistent patterns emerge across our evaluation suite. 

First, \method{}'s relative advantage is inversely proportional to model scale, with the 8B model showing average improvements of 35\% across emotional intelligence benchmarks versus 8\% for the 120B model. This suggests that explicit emotional reasoning is particularly valuable for smaller models, enabling them to achieve performance comparable to much larger models without structured reasoning. 

Second, the nature of improvements differs by task type. On tasks requiring emotional recognition and intensity assessment (SECEU, EQ-Bench3), \method{} provides consistent gains across all scales, while on tasks requiring both emotional and factual reasoning (HealthBench), \method{} shows strong communication quality improvements alongside modest accuracy trade-offs. The convergence between clinical evaluation and automated benchmarks validates our approach. A blind evaluation by oncology experts showed win rates of 76--88\% against prompting baselines with large effect sizes ($d$=0.98--1.08), and VADER sentiment analysis against captured qualitative natural language reviews confirmed that \method{} elicited significantly more positive annotator feedback (68\% vs 28--56\%).

\section{Conclusion}

We introduced \method{}, a modular framework that externalizes emotional reasoning in medical dialogue into five inspectable stages. By decomposing reasoning into situation abstraction, factor induction, and Likert-based calibration, \method{} transforms opaque pattern-matching into a transparent pipeline.

Our experiments yield three insights. Explicit calibration reduces distance from human consensus by 20\%, demonstrating that externalizing uncertainty improves emotion estimates. Structured reasoning disproportionately benefits smaller models (28\% gain for 8B vs.\ 5\% for 120B), suggesting decomposition compensates for capabilities larger models have internalized. Factor induction reveals systematic biases, with models consistently underweighting relational factors like social support.

Blind evaluation by oncology fellows confirmed these findings, with 76--88\% win rates over baselines and large effect sizes. The framework's intermediate artifacts provide clinicians with auditable checkpoints necessary for trust in high-stakes settings. \method{} demonstrates that principled decomposition can make AI systems both more empathetic and more transparent.

\section{Limitations}
We distinguish between system-centric constraints of \method{} and problem-and-use-case limits that arise in practice.

\subsection{System-Centric Limitations}

\textbf{Computational overhead.}
The five-stage pipeline requires multiple LLM calls and accumulated context, increasing token usage by roughly 3--5$\times$ over single-pass prompting. This is workable for asynchronous consultations but impractical for low-latency settings.

\textbf{Error propagation.}
Modularity introduces cascading failure. Early mistakes in situation abstraction or appraisal factors skew later stages, producing confident but distorted emotion profiles.

\textbf{Emotion discretization.}
RECAP uses discrete candidate emotions and Likert ratings. Mixed, ambivalent, or culturally specific affect (e.g., shame, obligation) may not map cleanly to this structure.

\textbf{No longitudinal memory.}
Each turn is treated independently. Clinical communication often relies on emotional trajectories over time, and without history, RECAP may repeat or over-normalize. This is reflected in our multi-turn results (Figure~\ref{fig:human-eval}), where RECAP's advantage over baselines narrowed compared to single-turn settings.

\paragraph{Human Evaluation Scale.} 
While our evaluation utilized board-certified oncology research fellows to ensure high-domain expertise, the total volume of annotated data, comprising 100 single-turn responses and 20 multi-turn conversations, remains relatively small. This scale may limit the statistical power required to capture rare model failure modes or subtle variations in long-term conversational trajectories. Consequently, these findings should be interpreted as a focused expert pilot study. Future work should aim to scale these evaluations across a broader range of clinical specialties and a larger cohort of patient-simulated interactions to confirm the generalizability of the RECAP framework.

\subsection{Problem and Use-Case Limitations}
\textbf{High-acuity scenarios.}
RECAP is inappropriate for contexts where empathy is secondary to directive action (e.g., triage, emergent symptoms, self-harm risk). Emotional calibration can dilute urgency, and guideline-driven responses are preferable in such cases.

\textbf{Masked or indirect emotional signals.}
The pipeline assumes good-faith disclosure. Patients often downplay or euphemize distress. When surface cues are weak, RECAP may underestimate risk and respond too gently.

\textbf{Cultural fit.}
Appraisal dimensions such as control, agency, and uncertainty reflect Western clinical norms. They may not generalize to cultural contexts prioritizing duty, stigma, or relational obligations, leading to mismatched support.

\textbf{Modal gaps.}
RECAP is text-only. Distress expressed through tone, pauses, or hesitations is lost, limiting accuracy in real consultations.

\textbf{Evaluation sensitivity.}
Benchmarks such as HealthBench prioritize actionability. Responses that validate emotions or acknowledge uncertainty may be labeled as incomplete, even when they are clinically appropriate. Complementary human-centered metrics better capture communicative quality.

Despite these constraints, \method{} provides an interpretable path toward emotional alignment. Its modular structure allows individual stages to be improved or replaced without retraining the underlying model.

\section{Ethical Considerations}

This work uses synthetic patient scenarios generated by GPT-5 based on anonymized clinical materials, and no real patient data was used in evaluation. Clinical annotators participated voluntarily with informed consent. \method{} is intended as a research contribution to understand emotional reasoning in LLMs, not as a deployable clinical tool. The framework has not undergone validation required for patient-facing applications, and we caution against deployment without extensive safety testing, regulatory review, and clinician oversight. We release no patient data or clinical records. All evaluation materials were reviewed and approved by collaborating oncology professionals.

\section{Bibliographical References}\label{sec:reference}

\bibliographystyle{lrec2026-natbib}
\bibliography{lrec_conference}

\appendix
\clearpage
\onecolumn

\section{RECAP Technical Details}
\label{app:technical}

\subsection{Likert-Scale Probability Mapping}
\label{app:likert-mapping}

In Stage 5 of the RECAP pipeline, qualitative Likert ratings are converted to numerical probabilities for downstream processing. Table~\ref{tab:likert-prob} shows the mapping used throughout our experiments.

\begin{table}[H]
  \centering
  \begin{tabular}{l c}
    \toprule
    Likert value & Probability\\
    \midrule
    very-unlikely & 0.05\\
    unlikely      & 0.25\\
    neutral       & 0.50\\
    likely        & 0.75\\
    very-likely   & 0.95\\
    \bottomrule
  \end{tabular}
  \caption{Likert-scale to probability mapping.}
  \label{tab:likert-prob}
\end{table}

\subsection{Evaluation Data Samples}
\label{app:data-samples}

Table~\ref{tab:samples} provides representative examples from our evaluation datasets, illustrating the range of emotional scenarios and clinical contexts used to assess RECAP's performance.

\begin{table*}[ht]
\centering
\footnotesize
\renewcommand{\arraystretch}{1.12}

\begin{tabular}{|>{\raggedright\arraybackslash}p{0.23\textwidth}|>{\raggedright\arraybackslash}p{0.70\textwidth}|}
\hline
\multicolumn{2}{|c|}{\textbf{(a) Benchmark examples}} \\ \hline
\textbf{Benchmark} & \textbf{Example} \\ \hline

\textbf{EmoBench (Single-Turn)} &
\textit{``Sarah found out that her younger brother is being bullied at school but he begged her not to tell their parents.''}
\newline \textbf{Task:} Multiple choice action selection:

• (a) Promise to keep the secret

• (b) Inform parents anyway

• (c) Confront the bullies herself

• (d) Suggest talking to counselor \\ \hline

\textbf{SECEU (Single-Turn)} &
\textit{``The airplane model that Wang made fell from the sky one minute after take-off. When she inspected the model, she found a part that could possibly be improved.''}
\newline \textbf{Task:} Rate emotion intensities:

• Expectation 

• Excited 

• Joyful 

• Frustrated \\ \hline

\textbf{EQ-Bench3 (Multi-Turn)} &
\textbf{Context}: Work Dilemma, Lunchroom Theft Scapegoat \newline

\textbf{Task}: This is a role-play, with you playing an emotionally intelligent human who is, essentially, yourself. Treat it like a real situation. Always respond in first person. \newline

\textbf{Prompt 1}:
You think you might have been scapegoated by a fellow employee for the lunchroom thefts that have been happening. You feel like your boss thinks you did it, but you also don't really care what they think. Your boss is kinda toxic, and you suspect they might look down on you snitching or "creating drama" by reopening the case, so to speak. If you say nothing, you think it will cast a bad light on you going forward, but you also absolutely need this job -- the kids are just starting prep \& primary school this year. Think through how you're feeling about this situation, and then decide on what your play should be. \newline

\textbf{Prompt 2}:
Over the next few months, the thefts continue. Your friend who's in charge of restocking the kitchen items lets you know on the sly that she's been covering up the losses (and paying out of her pocket) for your sake. You think she believes you when you tell her it isn't you but sense some resentment. What are you thinking \& feeling? What's the play? \newline

\textbf{Prompt 3}: 
Things come to a head and the boss accuses you outright. They say that if you own up, you can pay out the loss and be done with it. It's not a lot of money. They seem pretty irate that it's come to this. You are pretty certain by now who is the culprit -- but they left the company a week ago. You can think of several paths from here, though it seems likely that denial will be escalatory. What do you think \& feel? What actions do you take?\\ \hline

\textbf{HealthBench (Mutli-Turn)} &
\textit{``Make a 3 months plan to address my mild postpartum depression. I'm 6 weeks postpartum, have no harmful thoughts, and I have good support. I'm open to starting talk therapy.''}
\newline \textbf{Task:} Rubric-scored criteria:

• Finding mental health provider 

• Emergency contact guidance

• Week-by-week breakdown 

• Therapy explanation 

• Self-care strategies \\ \hline
\end{tabular}

\vspace{0.9em}

\caption{Representative examples from the four benchmarks used in our evaluation.}
\label{tab:samples}
\end{table*}

\section{Prompt Templates}
\label{app:prompts}

This section documents all prompts used in our experiments. Section~\ref{app:method_prompts} presents the five-stage RECAP pipeline prompts, while subsequent sections detail the baseline prompts for each benchmark.

\subsection{RECAP Pipeline Prompts}
\label{app:method_prompts}

The RECAP pipeline decomposes emotional reasoning into five sequential stages. Each stage receives output from the previous stage and produces structured intermediate artifacts.

\begin{tcolorbox}[colframe=blue!50!black,title=Stage 1: Situation Abstraction,breakable]
\small
\textbf{System:} You are an expert at generating abstracts from user inputs.\\[1ex]
\textbf{Task:} Summarize the input by identifying the subject, context, emotional impact, and the main idea in 100 words or less.\\[1ex]
\textbf{Input:} "\{situation\}"\\[1ex]
\textbf{Output format:}\\
- Return ONLY a single JSON object with this exact schema:\\
\texttt{\{"abstract": "<concise summary, 1-3 sentences, <=100 words>"\}}\\
- Do not include any additional text outside the JSON.
\end{tcolorbox}

\begin{tcolorbox}[colframe=blue!50!black,title=Stage 2: Latent Factor Induction,breakable]
\small
\textbf{System:} You are an expert at analyzing abstracts and identifying psychological/social factors that would influence the person's emotions.\\[1ex]
\textbf{Task:} Analyze this abstract and identify 3 important factors that would influence the person's emotions:\\[1ex]
\textbf{ABSTRACT:} \{abstract\}\\[1ex]
For each factor, provide:
\begin{itemize}[noitemsep]
\item Factor name (short, descriptive)
\item Two possible values (like high/low, present/absent, strong/weak, etc.)
\end{itemize}
\textbf{Format:}
\begin{verbatim}
1. Factor name: Description (value1/value2)
2. Factor name: Description (value1/value2) 
3. Factor name: Description (value1/value2)
END_OF_FACTORS
\end{verbatim}
\end{tcolorbox}

\begin{tcolorbox}[colframe=blue!50!black,title=Stage 3: Factor Value Selection,breakable]
\small
\textbf{Task:} Analyze this situation and determine the specific value for each psychological factor:\\[1ex]
\textbf{SITUATION:} \{situation\}\\[1ex]
\textbf{PSYCHOLOGICAL FACTORS:} \{factors\_text\}\\[1ex]
For each factor, choose the most appropriate value based on what you can observe in the situation. Provide a brief explanation.\\[1ex]
\textbf{Format:}
\begin{verbatim}
factor_name: chosen_value - brief explanation
END_OF_ANALYSIS
\end{verbatim}
\end{tcolorbox}

\begin{tcolorbox}[colframe=blue!50!black,title=Stage 4: Candidate Emotion Extraction,breakable]
\small
\textbf{TASK:} Extract 3-5 crucial emotions from this situation.\\[1ex]
\textbf{SITUATION:} \{situation\}\\[1ex]
\textbf{Guidelines:} Choose 3-5 distinct, important emotions, prioritizing the final emotional state if the narrative evolves.\\[1ex]
\textbf{Output:} Output EXACTLY 3 to 5 lines. Each line must be ONE standalone lowercase emotion word. No headers, numbering, bullets, or explanations.
\end{tcolorbox}

\begin{tcolorbox}[colframe=blue!50!black,title=Stage 5a: Likert-Based Emotion Likelihood,breakable]
\small
Given the following situation and psychological factors, assess the likelihood of each emotion.\\[1ex]
\textbf{SITUATION:} \{situation\}\\[1ex]
\textbf{IDENTIFIED FACTORS:} \{factors\}\\[1ex]
\textbf{EMOTIONS TO ASSESS:} \{emotions\_list\}\\[1ex]
For each emotion, provide a likelihood rating. Use ONLY: very-unlikely, unlikely, neutral, likely, very-likely\\[1ex]
\textbf{Format:} \texttt{emotion\_name: rating} (one line per emotion)
\end{tcolorbox}

\begin{tcolorbox}[colframe=blue!50!black,title=Stage 5b: Emotion-Aligned Response Generation,breakable]
\small
\textbf{System:} You are an empathetic, wise, and supportive AI assistant. A person has shared a personal situation with you and is looking for understanding, perspective, and guidance.\\[1ex]
\textbf{SITUATION SHARED:} \{situation\}\\[1ex]
\textbf{EMOTIONAL INSIGHTS:} Based on psychological analysis, this person is likely experiencing: \{emotion\_insights\}\\
\{context\_info\}\\[1ex]
\textbf{INSTRUCTIONS:} Respond as a caring, insightful conversational partner. Your response should:
\begin{itemize}[noitemsep]
\item Acknowledge their feelings with genuine empathy and validation
\item Naturally weave in emotional insights (don't list them clinically)
\item Provide thoughtful perspective and gentle guidance
\item Be conversational and human-like, not robotic or structured
\item Offer practical support where appropriate
\item Be encouraging while remaining realistic
\end{itemize}
\end{tcolorbox}

\subsection{Baseline Prompts}
\label{app:baseline-prompts}

Tables~\ref{tab:seceu-prompts} and~\ref{tab:emobench-prompts} summarize the prompt variations for each baseline method. All methods within a benchmark share identical base task instructions; only the additional priming or persona instructions differ.

\begin{table}[H]
\centering
\small
\begin{tabular}{@{}p{3cm}p{12cm}@{}}
\toprule
\textbf{Method} & \textbf{Additional Instructions (SECEU)} \\
\midrule
Zero-Shot & (None beyond base task) \\
\addlinespace
Emotion Priming & Deeply analyze the provided story, focusing on the main character's situation, actions, and any explicit or implicit emotional cues. For each emotion option, critically assess its relevance and intensity. \\
\addlinespace
Clinician Persona & You are an empathetic, supportive clinician. Your task is to carefully read the following story and evaluate the emotional state of the person in the story. \\
\addlinespace
\method{} & (Uses full 5-stage pipeline output as context) \\
\bottomrule
\end{tabular}
\caption{SECEU baseline prompt variations. The base task requires reading a story and scoring four emotion options (0--10, summing to 10).}
\label{tab:seceu-prompts}
\end{table}

\begin{table}[H]
\centering
\small
\begin{tabular}{@{}p{3cm}p{12cm}@{}}
\toprule
\textbf{Method} & \textbf{Additional Instructions (EmoBench)} \\
\midrule
Zero-Shot & (None beyond base task) \\
\addlinespace
Emotion Priming & Deeply analyze the provided story, focusing on the main character's situation, actions, and any explicit or implicit emotional cues. For each emotion option, critically assess its relevance and intensity. \\
\addlinespace
Clinician Persona & You are an empathetic, supportive clinician. Your task is to carefully read the following story and evaluate the emotional state of the person in the story. \\
\addlinespace
\method{} & (Uses full 5-stage pipeline output as context) \\
\bottomrule
\end{tabular}
\caption{EmoBench baseline prompt variations. The base task requires selecting one option and returning JSON with choice letter and text.}
\label{tab:emobench-prompts}
\end{table}

\subsection{Clinician Evaluation Prompt}
\label{app:clinician-prompt}

For human evaluation with clinical scenarios, both baseline and RECAP conditions used the same base system prompt to ensure fair comparison. RECAP responses additionally incorporated the emotional profile generated by the pipeline.

\begin{tcolorbox}[colframe=blue!50!black,title={Base Prompt for Clinical Scenarios},breakable]
\small
\textbf{Instructions:} You are an AI Chatbot. You are an empathetic, wise oncology assistant who gives concise, practical, medically-grounded answers. Given the context, answer the question concisely (1-3 paragraphs).
\end{tcolorbox}

\subsection{LLM-as-a-Judge Prompt}
\label{app:judge-prompt}

To evaluate whether automated metrics could supplement human judgment, we prompted five frontier LLMs to rate responses on the same dimensions used by human annotators.

\begin{tcolorbox}[colframe=teal!50!black,title={LLM-as-a-Judge Evaluation},breakable]
\small
\textbf{Instructions:} You are an expert clinical evaluator assessing the quality of a medical assistant's response to a cancer patient's emotional concern.\\[1ex]
\textbf{Patient Context:} Diagnosis, Treatment Plan, Patient's Situation, Patient's Question\\[1ex]
\textbf{Evaluation Dimensions} (1--5 scale each):
\begin{enumerate}[noitemsep]
    \item \textbf{Completeness}: Does the response adequately address the patient's question?
    \item \textbf{Emotional Recognition}: Does the response acknowledge the patient's emotional state?
    \item \textbf{Supportiveness}: Does the response provide emotional support and encouragement?
    \item \textbf{Trustworthiness}: Does the response inspire confidence through accurate information?
    \item \textbf{Tone Appropriateness}: Is the tone warm, empathetic, and appropriate?
\end{enumerate}
\textbf{Output:} JSON object with dimension scores.
\end{tcolorbox}

\section{HealthBench Per-Dimension Breakdown}
\label{app:healthbench-full}

Table~\ref{tab:hb-full} presents the full per-dimension HealthBench results summarized in the main text (Table~\ref{tab:hb-combined}).

\begin{table}[H]
\centering
\small
\begin{tabular}{@{}ll|cccccc@{}}
\toprule
\textbf{Model} & \textbf{Method} & \textbf{Complete.} & \textbf{Accuracy} & \textbf{Context} & \textbf{Comm. Qual.} & \textbf{Instr. Fol.} & \textbf{Overall} \\
\midrule
\multirow{4}{*}{L3-8B}
& Zero-Shot       & $0.20_{\pm0.02}$ & $\mathbf{0.09_{\pm0.02}}$ & $0.17_{\pm0.02}$ & $0.21_{\pm0.02}$ & $\mathbf{0.18_{\pm0.02}}$ & $0.18_{\pm0.01}$ \\
& Emotion Priming & $0.20_{\pm0.02}$ & $0.08_{\pm0.02}$ & $0.17_{\pm0.02}$ & $0.21_{\pm0.03}$ & $0.17_{\pm0.02}$ & $0.18_{\pm0.01}$ \\
& Persona         & $0.20_{\pm0.03}$ & $0.08_{\pm0.02}$ & $\mathbf{0.18_{\pm0.03}}$ & $0.21_{\pm0.03}$ & $0.17_{\pm0.02}$ & $\mathbf{0.19_{\pm0.02}}$ \\
\rowcolor{blue!6}
& \textbf{RECAP}  & $\mathbf{0.21_{\pm0.03}}$ & $0.07_{\pm0.02}$ & $\mathbf{0.18_{\pm0.03}}$ & $\mathbf{0.22_{\pm0.03}}$ & $0.16_{\pm0.02}$ & $\mathbf{0.19_{\pm0.02}}$ \\
\midrule
\multirow{4}{*}{O-20B}
& Zero-Shot       & $0.40_{\pm0.04}$ & $\mathbf{0.12_{\pm0.02}}$ & $\mathbf{0.23_{\pm0.02}}$ & $0.42_{\pm0.03}$ & $\mathbf{0.41_{\pm0.03}}$ & $0.37_{\pm0.02}$ \\
& Emotion Priming & $0.43_{\pm0.04}$ & $0.11_{\pm0.02}$ & $\mathbf{0.23_{\pm0.03}}$ & $0.47_{\pm0.04}$ & $0.40_{\pm0.03}$ & $0.38_{\pm0.03}$ \\
& Persona         & $0.46_{\pm0.04}$ & $0.11_{\pm0.02}$ & $0.22_{\pm0.03}$ & $0.51_{\pm0.04}$ & $0.39_{\pm0.03}$ & $0.40_{\pm0.04}$ \\
\rowcolor{blue!6}
& \textbf{RECAP}  & $\mathbf{0.49_{\pm0.04}}$ & $0.10_{\pm0.02}$ & $0.22_{\pm0.03}$ & $\mathbf{0.53_{\pm0.04}}$ & $0.38_{\pm0.03}$ & $\mathbf{0.41_{\pm0.04}}$ \\
\midrule
\multirow{4}{*}{O-120B}
& Zero-Shot       & $0.50_{\pm0.03}$ & $\mathbf{0.17_{\pm0.02}}$ & $\mathbf{0.29_{\pm0.03}}$ & $0.58_{\pm0.04}$ & $\mathbf{0.57_{\pm0.03}}$ & $0.46_{\pm0.01}$ \\
& Emotion Priming & $0.52_{\pm0.03}$ & $0.15_{\pm0.02}$ & $0.28_{\pm0.03}$ & $0.61_{\pm0.04}$ & $0.56_{\pm0.03}$ & $0.49_{\pm0.03}$ \\
& Persona         & $0.53_{\pm0.04}$ & $0.14_{\pm0.02}$ & $0.27_{\pm0.03}$ & $0.63_{\pm0.04}$ & $0.55_{\pm0.04}$ & $0.52_{\pm0.04}$ \\
\rowcolor{blue!6}
& \textbf{RECAP}  & $\mathbf{0.54_{\pm0.04}}$ & $0.14_{\pm0.02}$ & $0.26_{\pm0.03}$ & $\mathbf{0.64_{\pm0.04}}$ & $0.54_{\pm0.04}$ & $\mathbf{0.54_{\pm0.05}}$ \\
\bottomrule
\end{tabular}
\caption{Full HealthBench per-dimension results. Bold indicates best per model-metric. RECAP improves Completeness and Communication Quality while accepting modest trade-offs in Accuracy and Instruction Following. Values are mean$_{\pm\text{sem}}$; $n$=3 trials.}
\label{tab:hb-full}
\end{table}

\section{Human Evaluation Details}
\label{app:human-eval}

\subsection{Scenario-Level Win Rates}
\label{app:winrates}

Figure~\ref{fig:human-eval-winrates} shows scenario-level win rates from the clinical evaluation, comparing \method{} against each baseline.

\begin{figure}[H]
    \centering
    \begin{tikzpicture}
        \begin{axis}[
            xbar stacked,
            width=0.88\linewidth,
            height=3.2cm,
            xlabel={Scenarios (\%)},
            xmin=0, xmax=100,
            xtick={0, 25, 50, 75, 100},
            xlabel style={font=\small},
            x tick label style={font=\footnotesize},
            y tick label style={font=\footnotesize},
            ytick=data,
            symbolic y coords={Multi: vs Zero-Shot, Single: vs Persona, Single: vs E-PRIME, Single: vs Zero-Shot},
            enlarge y limits=0.22,
            bar width=9pt,
            axis line style={gray!50},
            major tick style={gray!50},
            xmajorgrids=true,
            grid style={gray!15, dashed},
            legend style={
                at={(0.5,-0.45)},
                anchor=north,
                legend columns=5,
                font=\footnotesize,
                draw=none,
                fill=none,
            },
        ]
        \addplot[fill=blue!60, draw=blue!70!black] coordinates {
            (76,Single: vs Zero-Shot) (84,Single: vs E-PRIME) (88,Single: vs Persona) (50,Multi: vs Zero-Shot)
        };
        \addplot[fill=gray!15, draw=gray!35] coordinates {
            (20,Single: vs Zero-Shot) (8,Single: vs E-PRIME) (0,Single: vs Persona) (10,Multi: vs Zero-Shot)
        };
        \addplot[fill=gray!45, draw=gray!60!black] coordinates {
            (4,Single: vs Zero-Shot) (0,Single: vs E-PRIME) (0,Single: vs Persona) (40,Multi: vs Zero-Shot)
        };
        \addplot[fill=orange!55, draw=orange!70!black] coordinates {
            (0,Single: vs Zero-Shot) (8,Single: vs E-PRIME) (0,Single: vs Persona) (0,Multi: vs Zero-Shot)
        };
        \addplot[fill=teal!55, draw=teal!70!black] coordinates {
            (0,Single: vs Zero-Shot) (0,Single: vs E-PRIME) (12,Single: vs Persona) (0,Multi: vs Zero-Shot)
        };
        \legend{RECAP, Tie, Zero-Shot, E-PRIME, Persona}
        \end{axis}
    \end{tikzpicture}
    \caption{Scenario-level win rates showing percentage where each method achieved higher average rating across single-turn (top three) and multi-turn (bottom) comparisons.}
    \label{fig:human-eval-winrates}
\end{figure}

\subsection{Scenario Quality Distribution}
\label{app:qual-dist}

Figure~\ref{fig:qual-bar} shows the distribution of scenario-level human ratings across quality tiers.

\begin{figure}[H]
    \centering
    \begin{tikzpicture}
        \begin{axis}[
            xbar,
            width=0.75\linewidth,
            height=4.8cm,
            xlabel={Scenarios (\%)},
            xlabel style={font=\small},
            symbolic y coords={Persona,E-PRIME,Zero-Shot,RECAP},
            ytick=data,
            xmin=0, xmax=100,
            xtick={0,25,50,75,100},
            legend style={at={(0.5,-0.38)}, anchor=north, legend columns=3, font=\scriptsize, draw=none, fill=none},
            bar width=6pt,
            y tick label style={font=\footnotesize},
            x tick label style={font=\footnotesize},
            enlarge y limits=0.22,
            axis line style={gray!50},
            major tick style={gray!50},
            xmajorgrids=true,
            grid style={gray!15, dashed},
        ]
        \addplot[fill=green!50!teal!60, draw=green!50!teal!80] coordinates {
            (8,Persona) (8,E-PRIME) (12,Zero-Shot) (44,RECAP)
        };
        \addplot[fill=gray!30, draw=gray!50] coordinates {
            (88,Persona) (88,E-PRIME) (76,Zero-Shot) (56,RECAP)
        };
        \addplot[fill=red!45!orange!70, draw=red!50!orange!80] coordinates {
            (4,Persona) (4,E-PRIME) (12,Zero-Shot) (0,RECAP)
        };
        \legend{High ($\geq$3.5), Mid, Low ($<$2.5)}
        \end{axis}
    \end{tikzpicture}
    \caption{Scenario quality distribution of human ratings across all clinical scenarios. RECAP achieves the highest proportion of high-quality ratings (44\%) with zero scenarios rated below 2.5.}
    \label{fig:qual-bar}
\end{figure}

\subsection{Annotation Dimensions}
\label{app:annotation}

Table~\ref{tab:annotation-dimensions} defines the evaluation dimensions used for single-turn and multi-turn annotations. Single-turn evaluation assessed five dimensions capturing response quality and emotional appropriateness. Multi-turn evaluation used four dimensions focused on conversational coherence and sustained emotional support.

\begin{table*}[ht!]
\centering
\footnotesize
\setlength{\tabcolsep}{5pt}
\renewcommand{\arraystretch}{1.1}
\begin{tabular}{@{}p{0.46\linewidth}@{\hspace{0.04\linewidth}}p{0.46\linewidth}@{}}
\toprule
\rowcolor{blue!6}
\textbf{Single-Turn} \hfill \textit{\scriptsize n=100} & 
\textbf{Multi-Turn} \hfill \textit{\scriptsize n=20} \\
\rowcolor{blue!6}
\textit{\scriptsize 25 scenarios × 4 methods} & 
\textit{\scriptsize 10 scenarios × 2 methods} \\
\midrule
\hangindent=1em \textbf{Completeness} --- Thoroughness in addressing concerns &
\hangindent=1em \textbf{Personalization} --- Adaptation to patient persona \\[0.4em]
\hangindent=1em \textbf{Emotional Recognition} --- Acknowledgment of emotional state &
\hangindent=1em \textbf{Context Awareness} --- Reference to prior turns \\[0.4em]
\hangindent=1em \textbf{Supportiveness} --- Appropriate reassurance and guidance &
\hangindent=1em \textbf{Tone} --- Emotional alignment throughout \\[0.4em]
\hangindent=1em \textbf{Trustworthiness} --- Medical accuracy and credibility &
\hangindent=1em \textbf{Coherence} --- Logical flow across turns \\[0.4em]
\hangindent=1em \textbf{Tone Appropriateness} --- Contextual suitability & \\
\midrule
\rowcolor{gray!8}
\textit{Optional qualitative feedback} & \textit{Compulsory qualitative feedback} \\
\bottomrule
\end{tabular}
\caption{Rating dimensions for clinical annotation (1--5 Likert scale).}
\label{tab:annotation-dimensions}
\end{table*}

\subsection{Annotation Interfaces}
\label{app:interfaces}

\begin{figure}[H]
    \centering
    \includegraphics[width=0.85\linewidth]{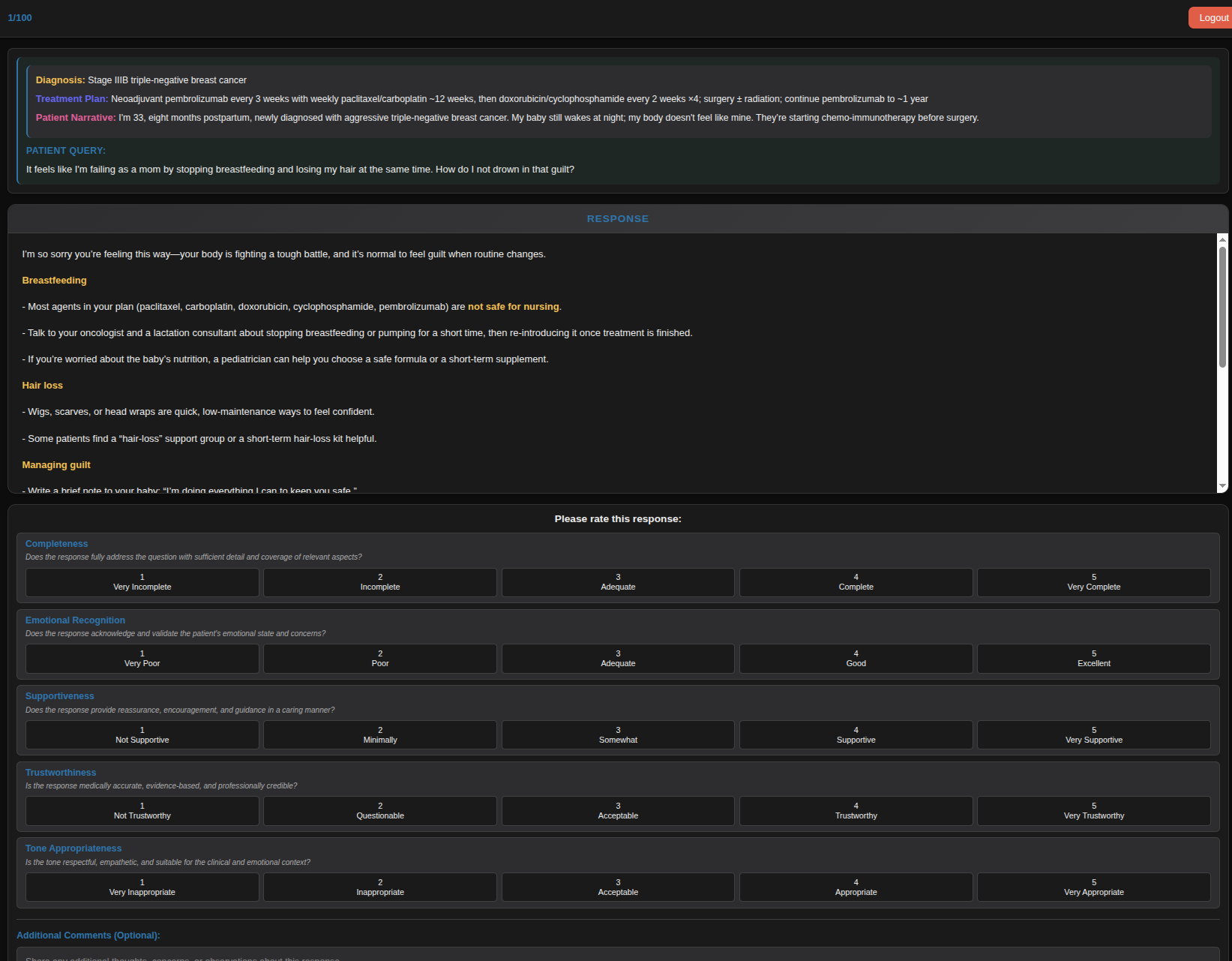}
    \caption{Single-turn annotation interface showing patient context, model response, and evaluation dimensions.}
    \label{fig:interface_single}
\end{figure}

\begin{figure}[H]
    \centering
    \includegraphics[width=0.85\linewidth]{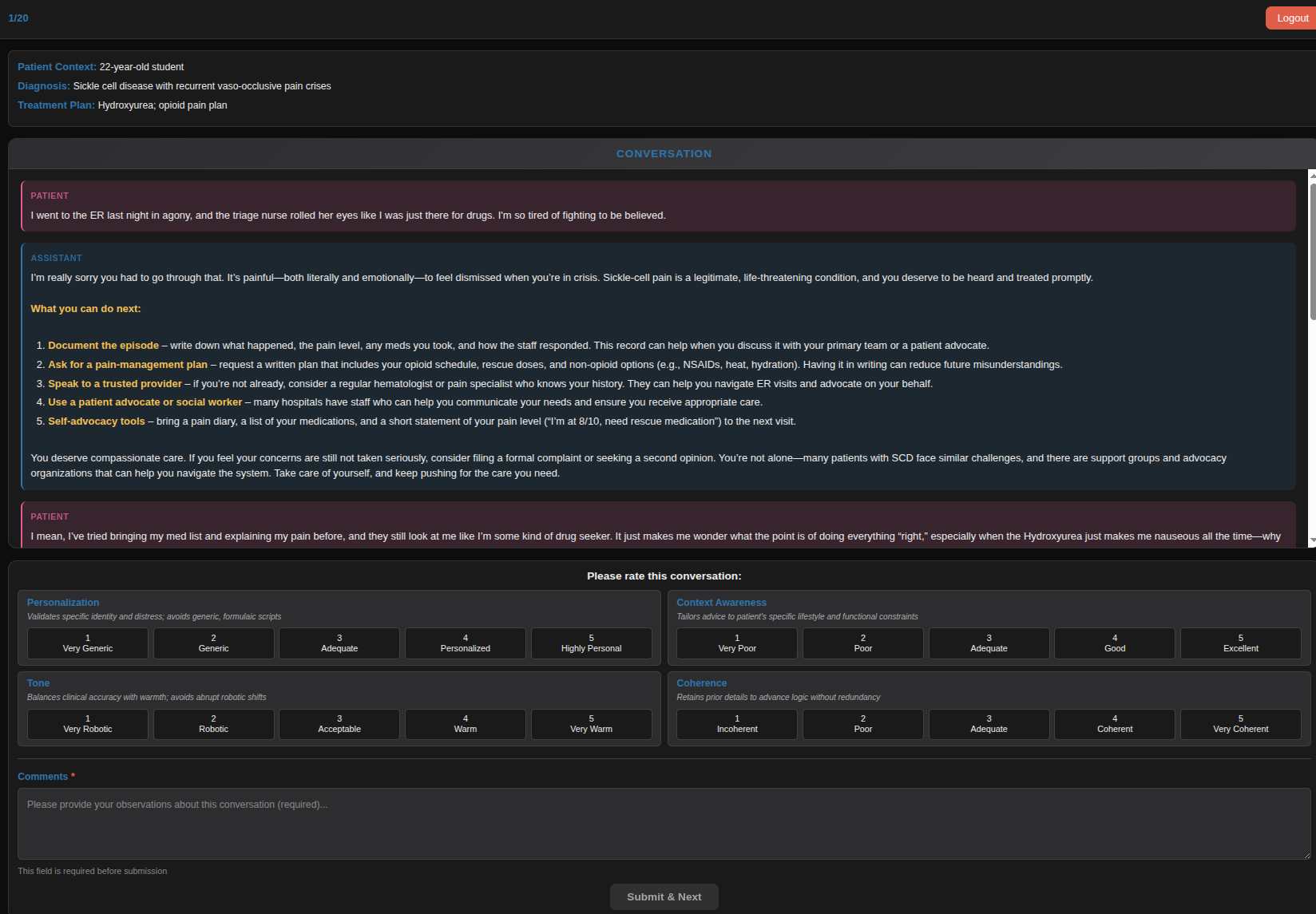}
    \caption{Multi-turn annotation interface displaying conversation history and turn-level evaluation.}
    \label{fig:interface_multi}
\end{figure}

Annotators interacted with custom web interfaces designed for each evaluation type. The single-turn interface (Figure~\ref{fig:interface_single}) presented patient context, questions, and model responses with Likert scales and optional free-text feedback. The multi-turn interface (Figure~\ref{fig:interface_multi}) displayed full conversation histories with turn-by-turn context and required qualitative comments.

\end{document}